\documentclass[twocolumn]{article}
\usepackage[utf8]{inputenc}
\usepackage{multicol}
\usepackage{booktabs}
\usepackage{tikz}
\title{Short-answer scoring with ensembles of pretrained language models}
\author{Christopher Ormerod \\ 
Cambium Assessment, Inc.\\
1000 Thomas Jefferson St., N.W.\\
Washington, D.C. 20007\\
christopher.ormerod@cambiumassessment.com}
\date{January 2022}

\begin{document}

\maketitle

\begin{abstract}
    We investigate the effectiveness of ensembles of pretrained transformer-based language models on short answer questions using the Kaggle Automated Short Answer Scoring dataset. We fine-tune a collection of popular small, base, and large pretrained transformer-based language models, and train one feature-base model on the dataset with the aim of testing ensembles of these models. We used an early stopping mechanism and hyperparameter optimization in training. We observe that generally that the larger models perform slightly better, however, they still fall short of state-of-the-art results one their own. Once we consider ensembles of models, there are ensembles of a number of large networks that do produce state-of-the-art results, however, these ensembles are too large to realistically be put in a production environment.
\end{abstract}


\section{Introduction}

Free-form constructed textual responses generally fall into one of two categories; essays and short answers. These two categories are not just distinguished by the average response length, they are also assessed very differently \cite{AESvsASAS}. Rubrics for essays often take grammatical rules, organization, and argumentation into consideration whereas rubrics for short answer questions tend to assess specific analytic or comprehension skills. This means that a response is not penalized if there are multiple spelling or grammatical errors present. Automated Short Answer Scoring (ASAS) and Automated Essay Scoring (AES) are two classes of techniques that utilize statistical models to approximate the assessment of constructed textual responses. Given the difference in rubrics, the performance of particular models and the importance of particular features used in each setting vary greatly. 

Traditionally, statistical models for AES have been based on bag-of-words (BoW) methods which combine frequency-based and hand-crafted features \cite{Erater1, IEA, PEG}. As neural networks were developed in other areas of NLP, they became increasingly adopted for AES \cite{AESNN1, AttAES, NNAES}. One of the most important developments in NLP has been the effectiveness of transformer-based pretrained language models such as the Bidirectional Encoder Representation by Transformers (BERT) model \cite{bert} which can be fine-tuned to a range of downstream tasks. The effectiveness of these models on the Kaggle essay dataset has been investigated by numerous authors \cite{efficient, OurBERT, bertfeatures}. More recently, we saw a combination of hand-crafted features and language models define the state-of-the-art on this dataset \cite{bertfeatures}.

The most effective methods for short answer questions differ depending on the various types of responses. In the case of the Powergrading dataset, where there are fewer than twenty words per response on average, a simple yet effective clustering technique is sufficient \cite{powergrading}. In the case of the SemEval-2013 Joint Student Response Analysis (SRA) task, the current state-of-the-art was achieved by fine-tuning BERT models \cite{semeval}. This work is concerned with the Kaggle Short Answer Scoring (KSAS) dataset\footnote{https://www.kaggle.com/c/asap-sas} \cite{kaggle2}. Each prompt consists of a passage and a prompt that asks the student to describe or explain aspects of the passage using evidence \cite{kaggle2}. Given the semantic nature of descriptions and explanations, we expect well-trained neural networks to perform well in this task. Despite the advancements of neural networks, the current state-of-the-art for this dataset has been achieved by the application of random forest classifiers to a set of rule-based features \cite{AutoSAS}. What is remarkable from a production standpoint is that the calculations for these models can be done in a low resource setting. 

Our goal of this short note is to explore how some of the most popular language models do when subjected to the KSAS dataset. We were expecting that language models on their own could surpass previous results, but when it comes to single models, on average, this is not the case. We are able to show that particular ensembles are capable of exceeding this benchmark, but the computational cost would be prohibitive from a production standpoint. In this sense, this work is the antithesis of \cite{AutoSAS} in that we simply bludgeon the problem to death with computational power. Even in doing so, there remain a few prompts that seem to fall drastically short of the methods in \cite{AutoSAS}. Conversely, there are some other prompts in which we see even our most efficient models perform comparably or even exceed rule-based methods, which we believe is sufficient to show that these methods and the results of this paper are of interest.

This paper is outlined as follows: In \S \ref{sec:method} we specify the way in which we have fine-tune, train, and ensemble the pretrained transformer-based language models and feature-based models, in \S \ref{sec:results} we present the results of the various models produced, and in \S \ref{sec:discussion} we discuss some corollaries of this work in terms of future directions. 

\section{Method}\label{sec:method}

Since BERT was introduced in \cite{bert}, a veritable cornucopia of language models have been introduced, each either varying the underlying architecture of BERT or the way in which they were trained. The General Language Understanding Evaluation (GLUE) benchmark has been one of several benchmarks used to evaluate the performance these language models in a range of classification, generation and understanding tasks \cite{glue}. We expect that some models, due to their architectural changes or training methods, should perform differently to other models. We shall compare each model by applying the same fine-tuning procedures to each pretrained model to each prompt in the KSAS dataset. 

We start by introducing the metrics typically used to evaluate model performance in most production systems in AES and ASAS \cite{williamson}. The primary statistic used in automated assessment is the Cohen's quadratic weighted kappa (QWK) score, defined by
\begin{equation}\label{eq:qwk}
\kappa = \frac{\sum \sum w_{ij} x_{ij}}{\sum \sum w_{ij} m_{ij}}
\end{equation}
where $x_{i,j}$ is the observed probability
\[
m_{i,j} = x_{ij}(1-x_{ij}),
\]
and
\[
w_{ij} = 1- \frac{(i-j)^2}{(k-1)^2},
\]
where $k$ is the number of classes. One interpretation of this statistic is that i represents the level of agreement between two scorers where you negate the agreement by chance. In production systems, we often require that the QWK between the true score and the scores predicted by the model are within 0.1 of the QWK between two humans. In an educational setting, most scoring engines are also required to have a standardized mean difference (SMD) with the final score of below 0.15 and the discrepancy between the IRR accuracy and the engines accuracy must be within some limit \cite{williamson}.

\begin{table}[]
    \centering
    \begin{small}
    \begin{tabular}{c|c c c | c c | c}
     & train & dev & test & \multicolumn{2}{c|}{dev} & Avg. \\
       Set & N & N & N & QWK & Acc & Length  \\ \toprule
        1 & 1337& 335 & 557 & 0.936 &  87.5\% & 52\\
        2 & 1022& 256 & 426 & 0.911 & 84.8\% & 65 \\
        3 & 1446& 362 & 406 & 0.758 & 78.5\% & 53\\
        4 & 1325& 332 & 295 & 0.686 & 78.3\% & 46\\
        5 & 1436& 359 & 598 & 0.935 & 95.9\% & 28\\
        6 & 1437& 360 & 599 & 0.951 & 97.0\% & 28\\
        7 & 1439& 360 & 599 & 0.973 & 96.4\% & 46 \\
        8 & 1439& 360 & 599 & 0.837 & 83.3\% & 60\\
        9 & 1438& 360 & 599 & 0.831 & 80.8\% & 54\\
        10 & 1312 & 328 &546 & 0.905 & 90.9\% & 45 \\
    \end{tabular}
    \end{small}
    \caption{The properties of the training and development set used in the training procedure.}
    \label{tab:IRRstats}
\end{table}

Our first step is to isolate a development set to be used for an early stopping mechanisms and hyperparameter-tuning. This set was chosen at random without any stratification. The properties of the development set are specified in Table \ref{tab:IRRstats}. Because the original test set has been withheld by the organizers of the competition, we use the public test set to validate the models. Unfortunately, this means that we do not have second reads for the validation set, hence, we cannot assess whether the results provided satisfies the operational criteria provided \cite{williamson}.

The training of a single model or model in an ensemble was performed using the AdamW optimizer \cite{adamw} with a linear learning rate scheduler. The loss function used was the usual binary-cross-entropy function. We train each model 20 epochs and select the model over that range the best QWK on the development set. To select the learning rate and the batch size, we used the Tree-structured Parzen Estimator (TPE) algorithm \cite{tpe} with 10 trials with batch sizes between 6 and 12 and learning rates between 5e-6 and 1e-4. We used the Optuna implementation of the TPE algorithm \cite{optuna}.  The source code we used to train and score for this project will be made available in a future version of this paper.

To keep the code accessible, we chose a range of models that were both popular, accessible through a single API\footnote{huggingface.co} and achieved high GLUE scores \cite{glue}. The selection of models, their references, and an approximation of their GLUE scores, and their respective sizes in millions of parameters is given in Table \ref{tab:bigguns}. 

\begin{table}[]
    \centering
    \begin{small}
    \begin{tabular}{c | l | c |c | c} \toprule
     Size & model & Ref. & Size & GLUE \\ \midrule 
     L &  ALBERT (xxL) & \cite{albert} & 222M &  89.4\\
      &  RoBERTa (large) & \cite{roberta} & 355M &  88.5\\
      &  BERT (large) & \cite{bert} & 335M &  88.1\\ 
      & Electra (large) & \cite{electra} & 335M & 89.4\\ \midrule
     B &  BERT (base) & \cite{bert} & 110M & 79.5 \\
      &  XLNet (base) & \cite{xlnet} & 117M & 79.6 \\
      &  RoBERTa (base) & \cite{roberta} & 124M & 79.6 \\ 
      &  Electra (base) & \cite{electra} & 110M & 82.7 \\ 
      & DeBERTa (base) & \cite{deberta} & 184M & \\\midrule
     S &  Electra (small) & \cite{electra} & 13M & 77.4 \\
      &  ConvBERT (med) & \cite{convbert} & 18M & 79.7 \\
      &  MobileBert & \cite{mobilebert} & 25M & 78.5 \\ 
      & Albert (base) & \cite{albert} & 17M & \\
      & DistBERT & \cite{distbert} & 67M & 77.0 \\\bottomrule
    \end{tabular}
    \end{small}
    \caption{A list of the various models used in this study in addition to the number of parameters and GLUE score. We have designated three different sizes as base }
    \label{tab:bigguns}
\end{table}

While each of these models are transformer-based pretrained language models, they differ in some key aspects. The RoBERTa models were trained longer and removes the next sentence prediction task in the original BERT \cite{roberta}. A novel aspect of the ALBERT models is the weight-sharing mechanism \cite{albert}. This mechanism was used in the base version to drastically reduce model size, and in the large version, to increase the size of the hidden units and feed-forward layers while keeping the size in terms of parameters managable. The Electra models differ greatly in the way they were trained; they are trained in generator and discriminator pair in which one model is used to generate tokens in masked positions, while the other attempts to distinguish between the generated and true labels \cite{electra}. The DeBERTa model we uses a disentangled attention mechanism in which the relative positions of words are considered, but not the absolute positions. Furthermore, the DeBERTa model is trained similarly to Electra however the goal of the discriminator in the case of DeBERTa is to detect replaced tokens rather than determine whether known tokens are generated or true \cite{deberta}.  

The difference between XLNet and BERT is that the tokens are essentially predicted simultaneously by considering all permutations of token prediction order \cite{xlnet}. The Convolutional BERT model is novel in that it replaces fully connected layers in the feed-forward mechanisms in BERT with convolutional layers that can be done more efficiently \cite{convbert}. The MobileBERT architecture uses linear layers called bottlenecks to reduce the dimension of the attention matrix computations, which is efficient and effective due to the rank of the attention matrix \cite{mobilebert}. Lastly, the Distilled BERT model has 6 layers instead of 12 and is trained using knowledge distillation. The authors claim this is a much smaller and faster version of BERT with 97\% of the performance \cite{distbert}. 

\begin{table*}
\begin{center}
\begin{tabular}{p{3cm} l  p{9cm}}
Name & $|v|$ & Description\\ \midrule
Sentence-BERT & 364 & The outut of the encoder component of Sentence-BERT.\\
TF-IDF & 100-300 & The transformation induced by the transformation of the largest 300 eigenvectors of the TF-IDF training matrix. \\
text-overlap & 15 & The number of minutia that intersect with the prompt.\\
key words, bigrams and trigrams & 90 & The key terms are extracted and near matches in the target text are counted.\\
text-stats & 10 & A number of key statistics like length, average word length, etc. \\ \bottomrule
\end{tabular}
\end{center}
\caption{A summary of the set of features we considered in the feature model.}\label{tab:features}
\end{table*}

Inspired by previous works both in essays and in short answer questions, we wanted to ensemble a range of models with a feature based model that captures the important components of the feature based models \cite{AutoSAS}. It has been our experience that ensembles between models of different natures better gains in performance than those that are similar. It is in this vein that we consider a feature model several key analogues of the features that were considered important. These features are summarized in Table \ref{tab:features}.

Firstly, we consider the embedding of documents delivered by the sentence embedding given by Sentence-BERT \cite{paraphrase}, a term-frequency inverse-document-frequency model, a set of text overlaps, and the frequencies of key $n$-grams for $n$ between $1$ and $3$. To evaluate text overlaps, we use minutia; any spaces, numbers, or punctuation are removed from the text and the prompt, at which point we count all overlapping strings between 5 and 20 characters, making a total of 15 dimensions. To evaluation "near matches" we use the text difference library difflib \footnote{https://docs.python.org/3/library/difflib.html}. There is a threshhold for the near matching cutoff that ranges from between 0.5, where half the characters are correct, and 1, in which case all the characters are correct. This threshold is varied in accordance with hyperparameter tuning choices. The reasons for this approach is that short answer scoring should disregard spelling, hence, the near-matches are a way to make the keyword features more robust to spelling errors. Lastly, we use a number of typical statistics concerned with word and sentence lengths. We normalize any features so that adhere to a normal distribution with mean 0 and a standard deviation of 1. 

Once all the features are compiled, we summarize the document into a single vector of featuers between dimension 579 and 779. A multilayer perceptron model is fit to the training set and the performance is then evaluated on the development set with some fixed learning rate and batch size. The learning rate, batch-size, TF-IDF dimension, and cutoff are subjected to a hyperparameter optimization with 20 trials. The goal of the feature model is to produce something with sufficient performance, that is distinct in nature, to ensemble with our pretrained language models.

\begin{figure}
    \centering
    \begin{tikzpicture}[xscale=1.35]
    
    \node[rounded corners=2pt, minimum width=1cm, draw=black, very thick, fill=green!5](t1) at (-2.5,1) {$tok_1$};
    \node[rounded corners=2pt, minimum width=1cm, draw=black, very thick, fill=green!5](t2) at (-1.5,1) {$tok_2$};
    \node[rounded corners=2pt, minimum width=1cm, draw=black, very thick, fill=green!5](t3) at (-.5,1) {$tok_3$};
    \node(tok2) at (1,1) {$\ldots$};
    \node[rounded corners=2pt, minimum width=1cm, draw=black, very thick, fill=green!5](tn) at (2.5,1) {$tok_n$};
    \node[rounded corners=2pt, minimum width=1cm, draw=black, very thick, fill=green!5](m1) at (-2.5,2) {$mod_1$};
    \node[rounded corners=2pt, minimum width=1cm, draw=black, very thick, fill=green!5](m2) at (-1.5,2) {$mod_2$};
    \node[rounded corners=2pt, minimum width=1cm, draw=black, very thick, fill=green!5](m3) at (-.5,2) {$mod_3$};
    \node(tok2) at (1,2) {$\ldots$};
    \node[rounded corners=2pt, minimum width=1cm, draw=black, very thick, fill=green!5](mn) at (2.5,2) {$mod_n$};
    \node[rounded corners=2pt, minimum width=1cm, draw=black, very thick, fill=green!5](l1) at (-2.5,3) {$lin_1$};
    \node[rounded corners=2pt, minimum width=1cm, draw=black, very thick, fill=green!5](l2) at (-1.5,3) {$lin_2$};
    \node[rounded corners=2pt, minimum width=1cm, draw=black, very thick, fill=green!5](l3) at (-.5,3) {$lin_3$};
    \node(tok2) at (1,2) {$\ldots$};
    \node[rounded corners=2pt, minimum width=1cm, draw=black, very thick, fill=green!5](ln) at (2.5,3) {$lin_n$};
    \node[rounded corners=2pt, minimum width=8cm, draw=black, very thick, fill=red!5](lr) at (0,4) {Logistic Regression};
    \node[rounded corners=2pt, draw=black, very thick, fill=red!5](score) at (0,5) {Score};
    
    \draw[->, very thick] (-2.5,0) -- (t1);
    \draw[->, very thick] (t1) -- (m1);
    \draw[->, very thick] (m1) -- (l1);
    \draw[->, very thick] (l1) -- (-2.5,3.7);
    \draw[->, very thick] (-1.5,0) -- (t2);
    \draw[->, very thick] (t2) -- (m2);
    \draw[->, very thick] (m2) -- (l2);
    \draw[->, very thick] (l2) -- (-1.5,3.7);
    \draw[->, very thick] (-.5,0) -- (t3);
    \draw[->, very thick] (t3) -- (m3);
    \draw[->, very thick] (m3) -- (l3);
    \draw[->, very thick] (l3) -- (-.5,3.7);
    \draw[->, very thick] (2.5,0) -- (tn);
    \draw[->, very thick] (tn) -- (mn);
    \draw[->, very thick] (mn) -- (ln);
    \draw[->, very thick] (ln) -- (2.5,3.7);
    \draw[->, very thick] (lr) -- (score);
    
    \node[rounded corners=2pt, minimum width=8cm, draw=black, very thick, fill=red!5](text) at (0,0) {Text};
    \end{tikzpicture}
    \caption{The structure of our general ensemble of models. The linear transformations, $lin_i$, are the outputs of the classification heads that give log-probabilities.}
    \label{fig:ensemble_struct}
\end{figure}
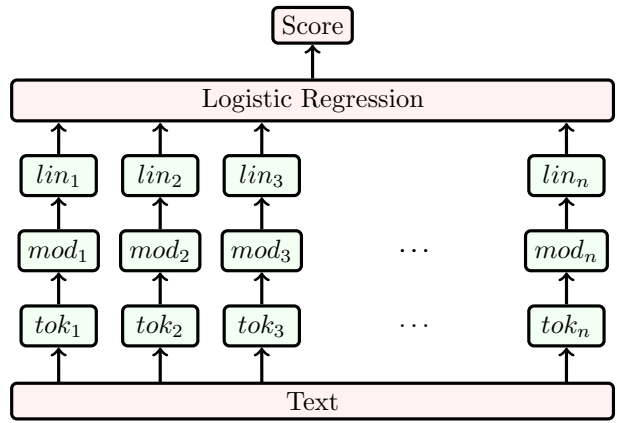

Given a selection of models, the structure of our ensemble is simple; we take the log-probabilities of each model on the development set as the training set of a logistic regression. The output of the logistic regression is considered to be the ensemble output. In this reigeme, the test set is only considered at this stage. The structure that we use for ensembling is depicted in Figure \ref{fig:ensemble_struct}.

\section{Results}\label{sec:results}

We optimized each model using a Xeon E5-2620 v4 @ 2.10GHz with a Nvidia RTX 8000 with 48Gb of on-board memory. This allowed for the batch sizes used for the larger models. Since the original test set for the competition has not been made available, we use the results on the publicly available test set, as was done for studies that are comparable to our own \cite{}. The individual models, and the ensemble results, are shown in Table \ref{tab:results}.

\begin{table*}[!ht]
    \centering
    \begin{footnotesize}
    \begin{tabular}{l|c|cccccccccc|c}
        && \multicolumn{10}{c}{QWK} & \\
        Model & ref. & 1 & 2 & 3 & 4 & 5 & 6 & 7 & 8 & 9 & 10 & mean\\ \hline
        
        Baseline & \cite{kaggle2} &  0.719 & 0.719 & 0.592 & 0.688 & 0.752 & 0.775 & 0.606 & 0.571 & 0.760 & 0.691 & 0.687\\
        Ramachandran et al. & \cite{Ramachandran} & 0.86 & 0.78 & 0.66 & 0.70 & 0.84 & 0.88 & 0.66 & 0.63 & 0.84 & 0.79 & 0.78\\
Riordan et al. & \cite{Riordan} & 0.795 & 0.718 & 0.684 & 0.700 & 0.830 & 0.790 & 0.648 & 0.554 & 0.777 & 0.735 & 0.723\\
        Kumar et al. & \cite{AutoSAS} & 0.872 & 0.824 & 0.745 & 0.743 & 0.845 & 0.858 & 0.725 & 0.624 & 0.843 & 0.832 & 0.791\\\midrule
        Features && 0.722& 0.664& 0.680& 0.712& 0.752& 0.776 & 0.624 & 0.514 & 0.789&  0.721 & 0.702\\\midrule
        AlBERT (xxL) & & 0.843 & 0.838 & 0.673 & 0.675 & 0.764 & 0.793 & 0.712 & 0.636 & 0.805 & 0.758 & 0.750 \\
        BERT(L) &&0.826 &0.791 & 0.715 & 0.732 & 0.762& 0.817 & 0.715 & 0.640 & 0.827 & 0.742 & 0.757\\
        Electra(L) \hfill 3 && 0.866 & 0.855 & 0.717 & 0.764 & 0.835 & 0.836 & 0.715 & 0.669 & 0.837 & 0.724 & 0.776\\
        RoBERTa (L) \hfill 2 & &  0.843 & 0.844 & 0.704 & 0.679 & 0.815 & 0.844 & 0.719 & 0.662 & 0.838 & 0.762 & 0.771 \\ \midrule
        BERT (base) && 0.849 & 0.772 & 0.692 & 0.722 & 0.845 & 0.840 & 0.676 & 0.598 & 0.829 & 0.717 & 0.749\\
        DeBERTa V3 (base) \hfill 1 && 0.881 & 0.865 & 0.677 & 0.690 & 0.766 & 0.830 & 0.722 & 0.690 & 0.840 & 0.733 & 0.769\\
        Electra (base) && 0.846 & 0.810 & 0.714 & 0.724 & 0.819 & 0.811 & 0.727 & 0.683 & 0.841 & 0.724 & 0.770\\
        RoBERTa (base) && 0.809 & 0.821 & 0.683 & 0.708 & 0.809 & 0.834 & 0.716 & 0.616 & 0.829 & 0.730 & 0.755\\
        XNet (base) && 0.822 & 0.784 & 0.675 & 0.680 & 0.814 & 0.824 & 0.686 & 0.663 & 0.804 & 0.718 & 0.747\\ \midrule
        ConvBERT & & 0.838 & 0.762 & 0.624 & 0.665 & 0.835 & 0.813 & 0.678 & 0.599 & 0.830 & 0.716 & 0.728\\
        DistilledBERT && 0.822 & 0.780 & 0.659 & 0.685 & 0.813 & 0.787 & 0.687 & 0.610 & 0.822 & 0.715 & 0.752\\
        Electra (small) & & 0.838 & 0.780 & 0.702 & 0.731 & 0.805 & 0.832 & 0.654 & 0.629 & 0.818 & 0.727 & 0.751\\
        MobileBERT & & 0.846 & 0.779 & 0.698 & 0.740 & 0.829 & 0.824 & 0.676 & 0.611 & 0.816 & 0.725 & 0.754\\
        \midrule
        Ensemble (best 2) && 0.884 & 0.868 & 0.714 & 0.728 & 0.790 & 0.839 & 0.730 & 0.690 & 0.850 & 0.761 & 0.786 \\
        Ensemble (best 3) && 0.882 & 0.891 & 0.722 & 0.750 & 0.813 & 0.822 & 0.734 & 0.702 & 0.865 & 0.779 & 0.796 \\
    \end{tabular}
    \end{footnotesize}
    \caption{The results of various small, base, and large language models. The three best models on the development set were indicated with 1,2, and 3 to the right along side the names. Ensemble 2 included the DeBERTa and the large RoBERTa, while the ensemble 3 included the large Electra models.}
    \label{tab:results}
\end{table*}

From a production standpoint, it is not just important that the QWK is high relative to the agreement between two human raters, it is also important that the SMD is within appropriate bounds \cite{williamson}. This is where the feature-based model does very well. In this situation, it is more important that the SMD is low across models, and we find that the ensembles do remarkably well. If we consider a typical violation to be a model with an SMD of over 0.15, most violations occur with small models. Generally speaking, as we found with QWK, the SMDs are far better for large models than the small models, but ensembles seem to do even better. Even ensembles of models with large SMDs have much better controlled SMDs than the individual models in the ensemble. If nothing else, this points to the fact that ensembles in the way we consider here give us a way of controlling SMDs. 

\begin{table*}[!ht]
    \centering
    \begin{footnotesize}
\begin{tabular}{lcccccccccc}
\toprule
{} &     1 &     2 &     3 &     4 &     5 &     6 &     7 &     8 &     9 &     10 \\
\midrule
Features       & 0.000 & 0.036 & 0.109 & 0.095 & 0.079 & 0.030 & 0.082 & 0.021 & 0.030 & 0.073 \\ \midrule

ALBERT        & 0.002 & 0.016 & 0.064 & 0.069 & 0.000 & 0.032 & 0.018 & 0.026 & 0.002 & 0.027 \\
BERT(L)       & 0.044 & 0.129 & 0.059 & 0.038 & 0.003 & 0.063 & 0.064 & 0.077 & 0.052 & 0.074 \\
Electra(L)    & 0.041 & 0.142 & 0.044 & 0.043 & 0.021 & 0.036 & 0.068 & 0.069 & 0.065 & 0.100 \\
RoBERTa (L)   & 0.025 & 0.077 & 0.097 & 0.208 & 0.029 & 0.089 & 0.043 & 0.014 & 0.015 & 0.086 \\ \midrule

BERT (base)         & 0.034 & 0.009 & 0.011 & 0.068 & 0.030 & 0.062 & 0.022 & 0.124 & 0.060 & 0.092 \\
DeBERTa V3 (base)  & 0.034 & 0.090 & 0.101 & 0.221 & 0.031 & 0.071 & 0.033 & 0.137 & 0.039 & 0.027 \\
Electra (base)  & 0.056 & 0.101 & 0.125 & 0.090 & 0.028 & 0.079 & 0.043 & 0.068 & 0.021 & 0.117 \\
RoBERTa (base)      & 0.141 & 0.132 & 0.079 & 0.156 & 0.039 & 0.007 & 0.078 & 0.047 & 0.050 & 0.073 \\
XNet (base)       & 0.080 & 0.082 & 0.145 & 0.021 & 0.036 & 0.107 & 0.064 & 0.122 & 0.019 & 0.211 \\ \midrule

ConvBERT      & 0.281 & 0.246 & 0.454 & 0.291 & 0.008 & 0.134 & 0.063 & 0.221 & 0.026 & 0.192 \\
DistilledBERT     & 0.023 & 0.048 & 0.102 & 0.016 & 0.028 & 0.050 & 0.104 & 0.016 & 0.037 & 0.211 \\
Electra (small)      & 0.057 & 0.158 & 0.081 & 0.057 & 0.070 & 0.055 & 0.103 & 0.117 & 0.019 & 0.099 \\
MobileBERT       & 0.023 & 0.091 & 0.045 & 0.072 & 0.060 & 0.044 & 0.060 & 0.106 & 0.060 & 0.069 \\\midrule
Ensemble (best of 2) & 0.004 & 0.062 & 0.011 & 0.093 & 0.037 & 0.010 & 0.024 & 0.002 & 0.024 & 0.047 \\
Ensemble (best of 3) & 0.016 & 0.065 & 0.004 & 0.082 & 0.045 & 0.022 & 0.012 & 0.008 & 0.017 & 0.081 \\
\bottomrule
\end{tabular}
\end{footnotesize}
    \caption{The SMD results of various small, base, and large language models used.}
    \label{tab:results}
\end{table*}

We find it interesting to note that there are a number of items in which pretrained models have succeeded where rule-based methods did not do as well. For example, several individual large pretrained language models exceed the state-of-the-art for prompts 2 and 8, yet pretrained models, or even their ensembles, seem to even come close to the results of \cite{AutoSAS} for item 10. It should be noted that even our naive feature model seemed to perform better than many of the pretrained language models tested. 

The three best models that performed on the development set were the large RoBERTa model, and the large and base Electra models in that order, even though that order is not reflected in the test set. When we ensemble the best two and three models that perform best on average, we obtain results that seem to be on-par and even slightly exceeding the state-of-the-art results of \cite{AutoSAS}. That said, the combination of these models is a model with approximately 875 million parameters. We do not believe it was a coincidence that the best models on both the test set and development set were among the largest, highest scoring models with respect to the GLUE benchmarks. 

\section{Discussion}\label{sec:discussion}

It seems to be the case that automated short answer scoring with pretrained transformer-based language models one their own can be outperformed generally by a mixture of regular expressions and other classical classifiers \cite{AutoSAS}. While we managed to exceed benchmarks with an ensemble of 3 very large networks, to do so with such huge computational power is a little bit dissatisfying. It is clear that such a solution is not feasible from a production standpoint. 

We firmly believe that the ideal solution, from a production and accuracy standpoint, would be the ensemble of an efficient network like \cite{efficient} and a rules based method like \cite{AutoSAS}. Firstly, this would require more careful consideration of the features used, and secondly, a careful consideration of how to incorporate these features into the score prediction. For example, concatenating the features to the set of features used by the classification head might yield better results \cite{bertfeatures}. We have generally found that ensembles 

In terms of architectures, there are a range of models we did not consider that would be worth mentioning. Some of these are  are a range of architectures that we did not consider, such as Reformer, LongerFormer, FNet, Linformer, Performer, MPNet. These all are variations on the transformer-based architecture that approximate attention using architectural differences that may be prove to be an advantage in short answer scoring. 

Lastly, we mention that there is still some work to be done in linking the output of the language model to the rubric. Most work on explainable AI has been focused on token-level importance, however, more semantically complex elements of a rubric are not simply stated in terms of the presence or absence of particular tokens. Knowing what features work well might be useful in determining interpretations of certain vectors in the feature space used to assign scores. This would be an important step in establishing a validity argument from these methods beyond their pure statistical performance.

\end{document}